\documentclass[conference]{IEEEtran}
\IEEEoverridecommandlockouts

\usepackage{cite}
\usepackage{amsmath,amssymb,amsfonts}
\usepackage{algorithmic}
\usepackage{graphicx}
\usepackage{textcomp}
\usepackage{hyperref}
\usepackage{tabularx}
\usepackage{booktabs}
\usepackage{xcolor}
\usepackage{tikz}
\usepackage{ragged2e}
\usepackage{float}
\usetikzlibrary{
    positioning,
    arrows.meta,
    calc
}
\usepackage{pifont}
\newcommand{\cmark}{\textcolor{green!60!black}{\ding{51}}}
\newcommand{\xmark}{\textcolor{red!70!black}{\ding{55}}}
\def\BibTeX{{\rm B\kern-.05em{\sc i\kern-.025em b}\kern-.08em
    T\kern-.1667em\lower.7ex\hbox{E}\kern-.125emX}}
\begin{document}

\title{Distilling 3D Spatial Reasoning into a Lightweight Vision-Language Model with CoT\\
\thanks{}
}


\author{
\IEEEauthorblockN{
Alaa Asfour, 
Christopher Indris,
Leihan Chen,
Tejas Vyas,
Guanghui Wang
}
\IEEEauthorblockA{Department of Computer Science, Toronto Metropolitan University, Toronto, ON, Canada}
\IEEEauthorblockA{Emails: \{alaa.asfour,wangcs\}@torontomu.ca}
}

\maketitle

\begin{abstract}
Large-scale 3D vision-language models (VLMs), such as LLaVA-3D, exhibit strong spatial reasoning capabilities but face significant deployment challenges due to their computational demands. We propose a knowledge distillation framework that transfers 3D spatial reasoning competence from a 7B-parameter teacher model to a 2.29B-parameter student model, achieving an 8.7$\times$ reduction in inference latency and a 3$\times$ reduction in model size while retaining 54-72\% of the teacher's performance on specialized spatial reasoning tasks. Our approach integrates VGGT (Visual Geometry Grounded Transformer) as the vision encoder and introduces a novel multi-task distillation pipeline with uncertainty-aware loss weighting. Distilled small models often lack the multi-step reasoning capacity of their large teachers when answering spatial queries. To address this without requiring chain-of-thought data or a CoT-capable teacher, we introduce Hidden Chain-of-Thought (CoT), a fixed set of learnable “thinking” tokens that function as an internal scratchpad before the final answer. To the best of our knowledge, this is the first application of latent scratchpad reasoning to distilled 3D VLMs from a teacher model (LLaVA-3D-7B), requiring no CoT-capable teacher or explicit chain-of-thought data. Within a unified architecture, the student model is trained to generate spatial descriptions, estimate depth, and detect objects. Central contributions include (i) spatial feature alignment across multi-view inputs, (ii) adaptive task loss weighting, and (iii) Hidden CoT for enhanced reasoning without altering the user-facing interface. Experimental evaluation on ScanNet and 3D-FRONT datasets demonstrates that the distilled model retains robust spatial relationship understanding, achieving 68-72\% accuracy in proximity and contact reasoning tasks, despite reduced text generation performance. The framework enables practical deployment of 3D VLMs on resource-constrained platforms while preserving the core spatial reasoning abilities required for robotics, augmented reality, and autonomous navigation applications. We position the work as teacher-relative indoor 3D scene QA, not embodied navigation. The source code is publicly available at the following \href{https://github.com/alaaasfour/distilled-LLaVA3D-with-CoT}{GitHub repository}.
\end{abstract}


\begin{figure*}[t]
\centering
\includegraphics[width=\textwidth]{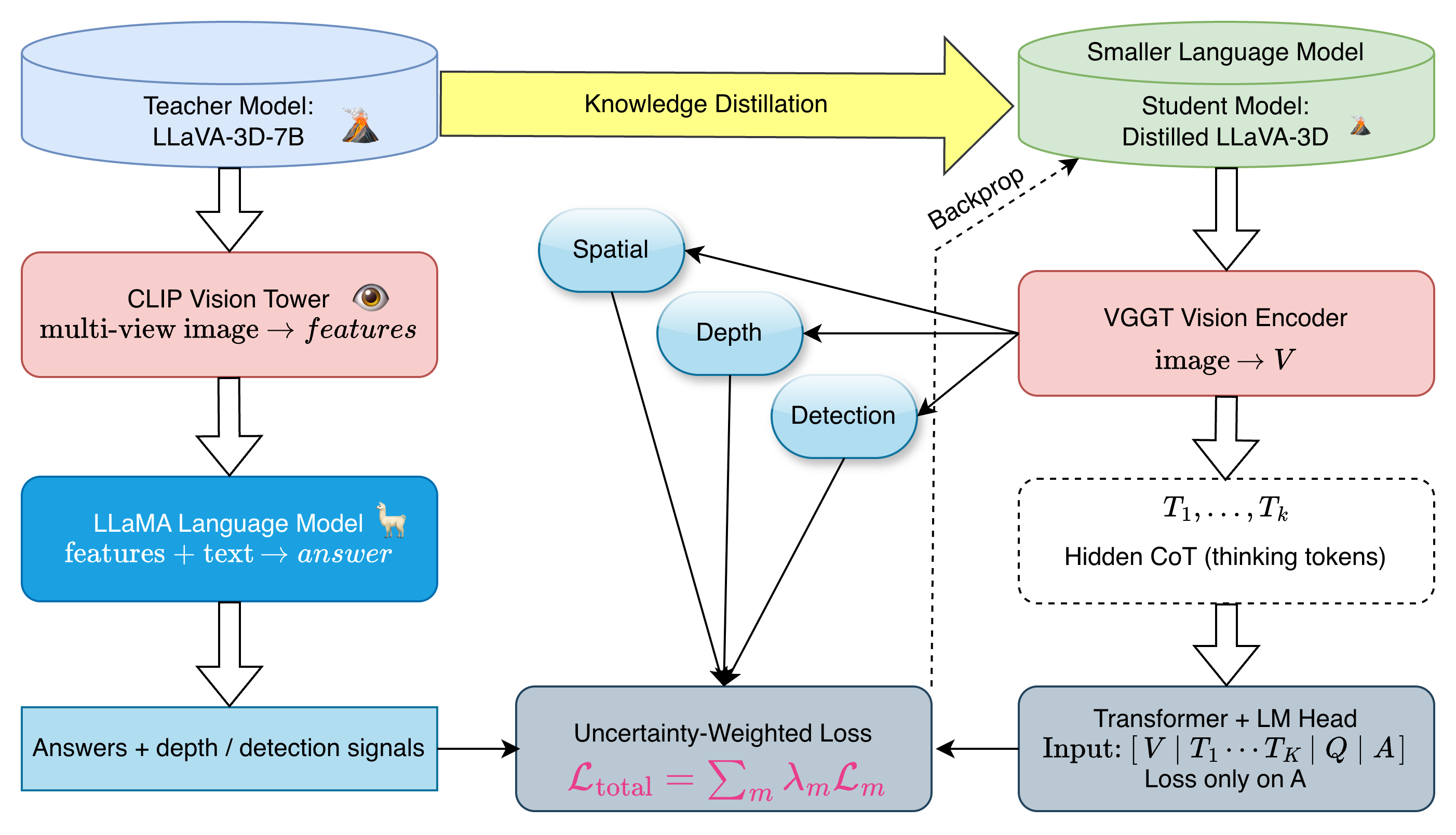}
\caption{Architecture and distillation pipeline. Left: Teacher (LLaVA-3D-7B) with CLIP vision tower and LLaMA provides answers and supervision. Center: Uncertainty-weighted combination of text, depth, detection, and spatial losses. Right: Student with VGGT vision encoder and a Hidden CoT block: $K$ learnable “thinking” tokens \(T_1..T_K\) sit between vision encoding \(V\) and question–answer \([Q][A]\); training and evaluation use only the final answer \(A\); \(T\) are never decoded. Multi-task heads (depth, detection, spatial) share the same vision features.}
\label{fig:fig_architecture}
\end{figure*}

\section{Introduction}
Vision-Language Models (VLMs) have revolutionized multimodal understanding by enabling machines to process and reason about visual and textual information simultaneously~\cite{radford2021learning}~\cite{alayrac2022flamingo}\cite{chen2024superlora}. Recent advances in 3D-aware VLMs, such as LLaVA-3D~\cite{zhu2024llava}, have extended these capabilities to three-dimensional spatial reasoning, enabling applications in robotics, augmented reality, and autonomous navigation~\cite{driess2023palm}~\cite{brohan2022rt}. However, the computational demands of these large-scale models limit their deployment in resource-constrained environments and real-time applications~\cite{strubell2019energy}\cite{chen2023mofa}. Although 3D VLMs demonstrate impressive performance on complex spatial reasoning tasks, practical deployment faces significant challenges: (1) \textbf{Computational overhead:} Large models require substantial GPU memory and inference time~\cite{hao2022learning}; (2) \textbf{Scalability:} The quadratic complexity of attention mechanisms in transformer-based architectures limits throughput~\cite{dao2022flashattention}; (3) \textbf{Resource constraints:} Edge devices and mobile platforms cannot accommodate models that exceed several gigabytes. These limitations prevent the widespread adoption of 3D VLMs in practical scenarios despite their superior capabilities.

Existing model compression techniques for 2D VLMs ~\cite{habib2023knowledge} have not been systematically applied to 3D-aware models. Although knowledge distillation has shown promise in reducing model size while preserving performance~\cite{hinton2015distilling}~\cite{gou2021knowledge}, its application to 3D spatial reasoning remains underexplored. Specifically, there is a lack of research on: (1) distilling 3D geometric understanding from large teacher models to compact student architectures; (2) integrating state-of-the-art vision encoders (e.g., VGGT~\cite{wang2025vggt}) into distilled 3D VLMs; (3) evaluating the trade-offs between efficiency and spatial reasoning accuracy in compressed 3D VLMs; and (4) improving reasoning in distilled 3D VLMs via latent scratchpad mechanisms (e.g., Hidden CoT) that require no chain-of-thought data or CoT-capable teacher. Our work addresses all four gaps and introduces Hidden CoT as a novel, interface-invariant upgrade for efficient 3D spatial reasoning.

This work presents a comprehensive study of knowledge distillation for 3D vision-language models, with the following contributions:
\begin{enumerate}
    \item \textbf{Architecture Design:} We propose a distilled 3D VLM architecture that employs VGGT as the vision encoder, achieving 3$\times$ model compression while preserving competitive spatial reasoning capabilities on indoor protocols.
    \item \textbf{Distillation Framework:} We develop a multi-task distillation pipeline that transfers knowledge across text generation, depth estimation, and object detection tasks, enabling the student model to learn rich 3D representations from the teacher model.
    \item \textbf{Efficiency Analysis:} We demonstrate significant efficiency gains: 8.7$\times$ increase in inference throughput on H-100 class GPUs and 3$\times$ reduction in model size compared to the teacher model (LLaVA-3D-7B), making deployment feasible on resource-constrained devices.
    \item \textbf{Comprehensive Evaluation:} We conduct a comprehensive quantitative analysis using standard metrics (BLEU, ROUGE, METEOR) and efficiency benchmarks, establishing baselines for future research in 3D VLM compression.
    \item \textbf{Hidden Chain-of-Thought (Scratchpad):} We propose Hidden CoT, a novel mechanism for distilled 3D VLMs, in which $K$ learnable “thinking” tokens are inserted between the vision encoding and the question-answer sequence. The model is trained and evaluated exclusively on the final answer; the scratchpad is never decoded or exposed to inference. To the best of our knowledge, this is the first latent scratchpad design for distilled 3D vision-language models, requiring no CoT supervision or CoT-capable teacher, and it improves reasoning while keeping the user interface and evaluation protocol unchanged, providing a distinctive and efficient framework for 3D spatial reasoning.
\end{enumerate}

\section{Related Work}
\paragraph{Vision-Language Models}
Vision-Language Models have evolved from simple image-captioning systems to sophisticated multimodal reasoning frameworks. Early approaches like CLIP demonstrated the power of contrastive learning for aligning visual and textual representations~\cite{radford2021learning}\cite{li2025pf3det}. Subsequent models such as LLaVA~\cite{liu2023visual} and BLIP~\cite{li2022blip} integrated large language models with vision encoders. These models typically employ transformer architectures with cross-modal attention mechanisms, achieving state-of-the-art performance on benchmarks like VQA~\cite{antol2015vqa} and GQA~\cite{hudson2019gqa}. However, most existing VLMs focus on 2D image understanding, leaving 3D spatial reasoning largely unexplored.

\paragraph{3D-Aware Multimodal Reasoning}
The extension of VLMs to 3D understanding has gained traction with models like 3D-LLM~\cite{hong20233d} and LLaVA-3D~\cite{zhu2024llava}. These models process multi-view images or point clouds to reason about spatial relationships, object locations, and geometric properties. LLaVA-3D, in particular, combines a vision tower with a language model to answer questions about 3D scenes, achieving strong results on spatial reasoning benchmarks. However, these models inherit the computational overhead of large-scale architectures, limiting their practical applicability. Our work addresses this limitation by distilling 3D reasoning capabilities into more efficient architectures.

\paragraph{Model Compression \& Distillation}
Knowledge distillation, introduced by Hinton et al.~\cite{hinton2015distilling}, has become a cornerstone of model compression. The technique transfers knowledge from a large teacher model to a smaller student model through soft target supervision. Recent advances have extended distillation to multimodal settings~\cite{zhao2024distilling}, showing that student models can achieve 80-90\% of teacher performance with 5-10$\times$ compression ratios. However, most distillation studies focus on 2D vision tasks, with limited exploration of 3D spatial reasoning. Our work bridges this gap by applying distillation to 3D VLMs while integrating modern vision encoders like VGGT~\cite{wang2025vggt}. We adopt VGGT for its strong geometric priors (e.g., camera and depth estimation) that directly benefit 3D scene understanding; although VGGT itself has on the order of 1B parameters, integrating it into our student pipeline yields improved spatial reasoning while keeping the full student model compact, smaller than the teacher (2.29B parameters total).

\begin{figure}[t]
\centering
\includegraphics[width=\columnwidth]{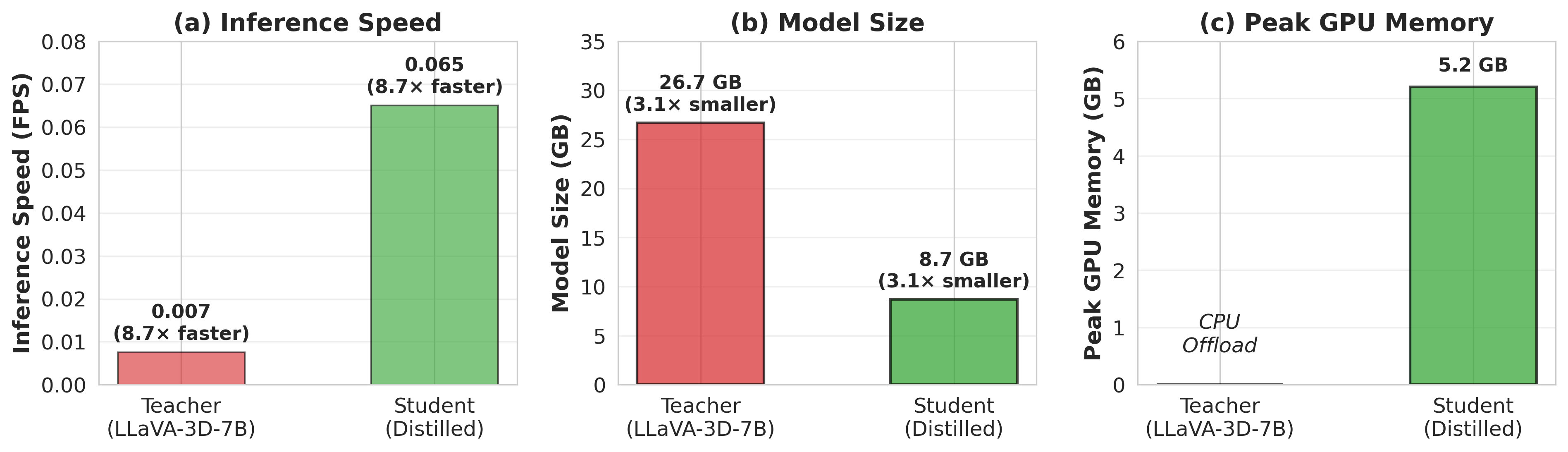}
\caption{Efficiency comparison between teacher and student models. (a) The student achieves 8.7$\times$ faster inference (0.065 FPS vs 0.0075 FPS). (b) Model size is reduced by 3$\times$ (8.7 GB vs 26.7 GB). (c) Peak GPU memory usage is 5.2 GB, making deployment feasible on consumer GPUs. The teacher model uses CPU offloading to manage memory constraints.}
\label{fig:efficiency}
\end{figure}

\paragraph{Chain-of-Thought (CoT) and Latent Reasoning}
CoT prompting~\cite{wei2022chain} and zero-shot CoT~\cite{kojima2022large} improve reasoning in large language models by eliciting explicit intermediate steps. These methods require either CoT-annotated data or a teacher model that can produce step-by-step reasoning when prompted. In contrast, \textit{latent} or \textit{hidden} reasoning uses internal representations that are never decoded to text: e.g., Quiet-STaR and internal-monologue-style models reason in hidden space. Scratchpad and schema-based networks use internal working memory to improve outputs. Our Hidden CoT differs from prior work in three ways: (1) we apply a fixed-length learnable scratchpad (K thinking tokens) to a distilled 3D VLM from LLaVA-3D, not to a general-purpose LLM; (2) we require no CoT data or CoT-capable teacher, training uses only the teacher’s final answers; (3) the design is interface- and evaluation-invariant, so deployment and benchmarks remain unchanged. To our knowledge, no prior work combines latent scratchpad reasoning with 3D VLM distillation in this way. This combination of teacher-relative efficiency, 3D-specific distillation, and Hidden CoT for improved reasoning advances scalable prototyping and analysis of compressed 3D VLMs, subject to indoor benchmark and throughput limitations.

\paragraph{Spatial Reasoning Benchmarks}
Evaluating 3D spatial reasoning requires specialized benchmarks that test geometric understanding, object relationships, and spatial queries. Datasets like ScanQA \cite{azuma2022scanqa}, 3D-SPS~\cite{luo20223d}, and EmbodiedScan~\cite{wang2024embodiedscan} provide standardized evaluation protocols for 3D VLMs. These benchmarks assess capabilities, including spatial localization, object detection, depth estimation, and natural language question answering about 3D scenes. Our evaluation framework incorporates metrics from these benchmarks while providing efficiency analysis crucial for practical deployment.

\section{Methodology}
\subsection{Overview of the Approach}
Our distillation framework consists of three main components: (1) a large teacher model (LLaVA-3D-7B)\cite{zhu2024llava} that provides supervision signals, (2) a compact student model with VGGT vision encoder\cite{wang2025vggt}, and (3) a multi-task distillation pipeline that transfers knowledge across text generation, depth estimation, and object detection tasks. The student model learns to mimic the teacher's behavior while maintaining significantly reduced computational requirements. Figure~\ref{fig:fig_architecture} illustrates the overall architecture and distillation pipeline.

\subsection{Teacher Model}
We employ LLaVA-3D-7B~\cite{zhu2024llava} as the teacher model, a state-of-the-art 3D VLM with 7 billion parameters. The model combines a vision tower (CLIP-based)~\cite{radford2021learning} with a language model (LLaMA)~\cite{touvron2023llama} to process multi-view images and generate textual responses about 3D scenes. The teacher model demonstrates strong performance on spatial reasoning tasks, achieving near-perfect scores on validation sets. During distillation, the teacher generates soft targets for text generation and provides supervision for depth estimation and object detection tasks.

\subsection{Student Model Architecture}
Our student model architecture consists of the following key components:

\textbf{Vision Encoder:} We integrate VGGT~\cite{wang2025vggt} as the vision encoder, replacing the teacher's CLIP-based vision tower. VGGT is specifically designed for 3D scene understanding, leveraging geometric priors and spatial attention mechanisms. The encoder processes input images at 518×518 resolution and outputs feature representations of dimension 2048, which are then projected to match the language model's embedding space.

\textbf{Language Model:} We employ a custom transformer-based language model architecture (2.29B parameters) compared to the teacher's 7B parameters. The architecture comprises 24 transformer layers, each with hidden size 2048, 16 attention heads, and a feedforward (MLP) intermediate dimension of 8192 (i.e., the inner dimension of the two-layer feedforward block in each layer). Unlike the teacher model, which uses LLaMA, our student model uses a custom transformer encoder initialized from scratch, enabling fine-grained control over model capacity and efficiency. The model uses cross-modal attention to integrate visual features from VGGT with textual inputs, allowing joint reasoning over visual and linguistic representations.

\textbf{Multi-task Heads:} To align with the teacher's multi-faceted supervision and to reinforce 3D spatial grounding beyond text alone, the student uses several task-specific heads in addition to language modeling. This design allows depth and detection to act as auxiliary objectives that improve the shared representation for spatial reasoning. The student model includes task-specific heads for: (1) Text Generation: Autoregressive language modeling for generating responses to spatial queries; (2) Depth Estimation: Regression head for predicting depth maps; (3) Object Detection: Detection head for localizing objects in 3D space.

\textbf{Hidden CoT (Scratchpad):} Inspired by latent (hidden) reasoning and internal scratchpad designs in language models~\cite{zelikman2403quiet}~\cite{wei2022chain}, to strengthen reasoning without changing the user-facing output, we add a Hidden CoT mechanism in the form of $K$ learnable "thinking" token embeddings (default $K=8$) inserted between the vision prefix and the question–answer sequence in the transformer.

(1) Sequence format: The input to the transformer is \([V] [T_1,\ldots,T_K] [Q] [A]\), where \(V\) is the projected vision (and depth) encoding (one token), \(T_i\) are the thinking tokens (fixed size, never decoded to text), \(Q\) is the question token sequence, and \(A\) is the answer token sequence from the teacher.
(2) Parameters: The thinking tokens are implemented as a learnable embedding matrix of shape \((1, K, \text{hidden\_size})\), initialized with small random values (e.g. \(\mathcal{N}(0, 0.02^2)\)), and broadcast to batch size in the forward pass. They receive no direct supervision; gradients flow only from the answer loss.
(3) Training objective: We apply causal language modeling (cross-entropy) only over the answer positions \(A\). Positions corresponding to \(V\), \(T\), and \(Q\) are masked out in the loss (e.g. label index \(-100\)). Thus, the thinking tokens are trained solely through backpropagation from the final-answer loss, encouraging them to capture latent structure that improves answer quality.
(4) Inference: At test time, we run the same architecture with input \([V][T_1..T_K][Q]\) and autoregressively generate only the answer tokens \(A\). The thinking tokens remain in the context but are never decoded or shown to the user; the interface and evaluation protocol (BLEU, ROUGE, spatial benchmarks) are unchanged.
(5) Integration with distillation: Hidden CoT is trained with the same teacher answers as the non-CoT student; no chain-of-thought data or "think step by step" teacher prompts are required. Auxiliary losses (depth, detection, spatial) are combined with the answer-only text loss via uncertainty-based weighting as in the baseline.

\textbf{Novelty} (1) To the best of our knowledge, Hidden CoT is the first latent scratchpad mechanism applied to distilled 3D vision-language models, where the student must compress both 3D spatial understanding and reasoning into a small footprint. (2) In contrast to explicit CoT~\cite{wei2022chain}, our model does not require step-by-step reasoning data or a teacher model that outputs intermediate rationales; the scratchpad is fully latent and learned end-to-end from final-answer supervision. (3) Deployment and evaluation remain unchanged; only the final prediction is ever shown or scored, making the method a drop-in upgrade for existing distilled 3D VLM pipelines. (4) The use of a small, fixed number of thinking tokens introduces only a marginal increase in parameters and latency, preserving the 8.7$\times$ speedup and 3$\times$ compression of the distilled model while offering a principled path to better spatial reasoning.

The complete student model contains 2.29 billion parameters (8.7 GB), achieving a 3$\times$ compression ratio compared to the teacher's 7 billion parameters (26.7 GB). When using the reduced "tiny" transformer configuration for 10GB GPUs (12 layers, 1024 hidden size), the model remains fully compatible with the same Hidden CoT design using smaller $K$ and shorter sequence lengths.

\subsection{Distillation Pipeline}
Our distillation pipeline employs a multi-task loss function that combines text generation, depth estimation, object detection, and spatial reasoning losses. The key innovation is our \textbf{Spatial Corresponding Distillation Loss}, which aligns spatial feature representations across multi-view inputs and ensures consistent 3D understanding.

\subsubsection{Text Generation Loss}
We use cross-entropy loss between the student and teacher predictions (temperature-scaled~\cite{hinton2015distilling}). Hidden CoT variant: For the sequence \([V][T_1..T_K][Q][A]\), we compute loss only over the answer tokens \(A\); positions for \(V\), \(T\), and \(Q\) are masked (e.g., label \(-100\)). The thinking tokens receive no direct supervision and are trained solely from the answer loss. This design adds latent reasoning without chain-of-thought data or evaluation changes.

\subsubsection{Auxiliary Losses (Depth, Detection, Spatial)}
We use: depth-L1 regression, cross-entropy for depth bins, and KL on depth distributions; detection-focal loss for class probabilities plus localization; spatial-feature L2 alignment with the teacher and cross-view consistency, plus KL on left/right and above/below distributions; multi-view-L2 consistency of student predictions across views; feature-alignment of detector probabilities, spatial distributions, and depth hints. All are combined with the text loss via uncertainty weighting.

\subsubsection{Spatial Corresponding Distillation Loss} 
The loss ensures spatial feature alignment across multi-view inputs by aligning spatial feature maps from the vision encoder with corresponding features from the teacher, enabling the student to learn consistent 3D representations:
\begin{equation}
\begin{aligned}
\mathcal{L}_{\text{spatial}} 
&= \sum_{v,i,j} \bigl\lVert F_s^{(v)}(i,j) - F_t^{(v)}(i,j) \bigr\rVert_2^2 \\
&\quad + \lambda_{\text{cross}} 
\sum_{v_1 \neq v_2} 
\operatorname{sim}\!\bigl(F_s^{(v_1)}, F_s^{(v_2)}\bigr)
\end{aligned}
\end{equation}
where \(F_s^{(v)}, F_t^{(v)}\) are student and teacher spatial features for view \(v\); the second term enforces cross-view consistency.

\subsubsection{Uncertainty-Based Total Loss}
We balance all task losses with learnable uncertainties \(\sigma_i\)~\cite{kendall2018multi}
\begin{equation}
\mathcal{L}_{total} = \sum_{i} \frac{1}{2\sigma_i^2} \mathcal{L}_i + \log \sigma_i
\end{equation}
where \(i\) runs over text, depth (CE, reg, KL), detection, spatial, multiview, and feature. This adaptively down-weights harder tasks and avoids manual loss tuning.

\subsection{Training Details}
We train the student model on a diverse dataset comprising ScanNet~\cite{dai2017scannet} and 3D-FRONT~\cite{huang2025midi}, totaling more than 2000 samples. The dataset is split into 80\% training samples and 20\% validation samples. Training is conducted using the following hyperparameters.
\textbf{Optimizer:} AdamW with a learning rate of $1 \times 10^{-4}$
\textbf{Batch size:} 1 (due to memory constraints)
\textbf{Epochs:} 2 (with early stopping configured but not triggered)
\textbf{Learning rate schedule:} Cosine annealing with warmup
\textbf{Loss weights:} Uncertainty-based adaptive weighting (learnable parameters)

To manage memory constraints, we employ CPU offloading for the teacher model. The uncertainty-based loss weighting automatically adjusts task importance during training, with weights adapting based on task difficulty and training progress. The baseline (non-CoT) training shows consistent learning, with training loss decreasing from 6.37 (epoch 1) to 6.00 (epoch 4) and validation loss improving from 0.873 (epoch 1) to 0.869 (epoch 3). Hidden CoT training (\(K=8\) thinking tokens, 2 GPUs) was run for 5 full epochs. Best validation loss was \textbf{4.79} at epoch \textbf{4}; training loss decreased from 7.93 (epoch 1) to 7.39 (epoch 5), representing a 6.8\% reduction. Task weights in the CoT run follow the same pattern: depth regression and detection receive the highest weights ($\sim$100–450 and $\sim$90–160), text stays low ($\sim$1.1–1.2), and spatial/multiview/feature remain moderate or fixed ($\sim$2.5).

\begin{figure}[t]
\centering
\includegraphics[width=\columnwidth]{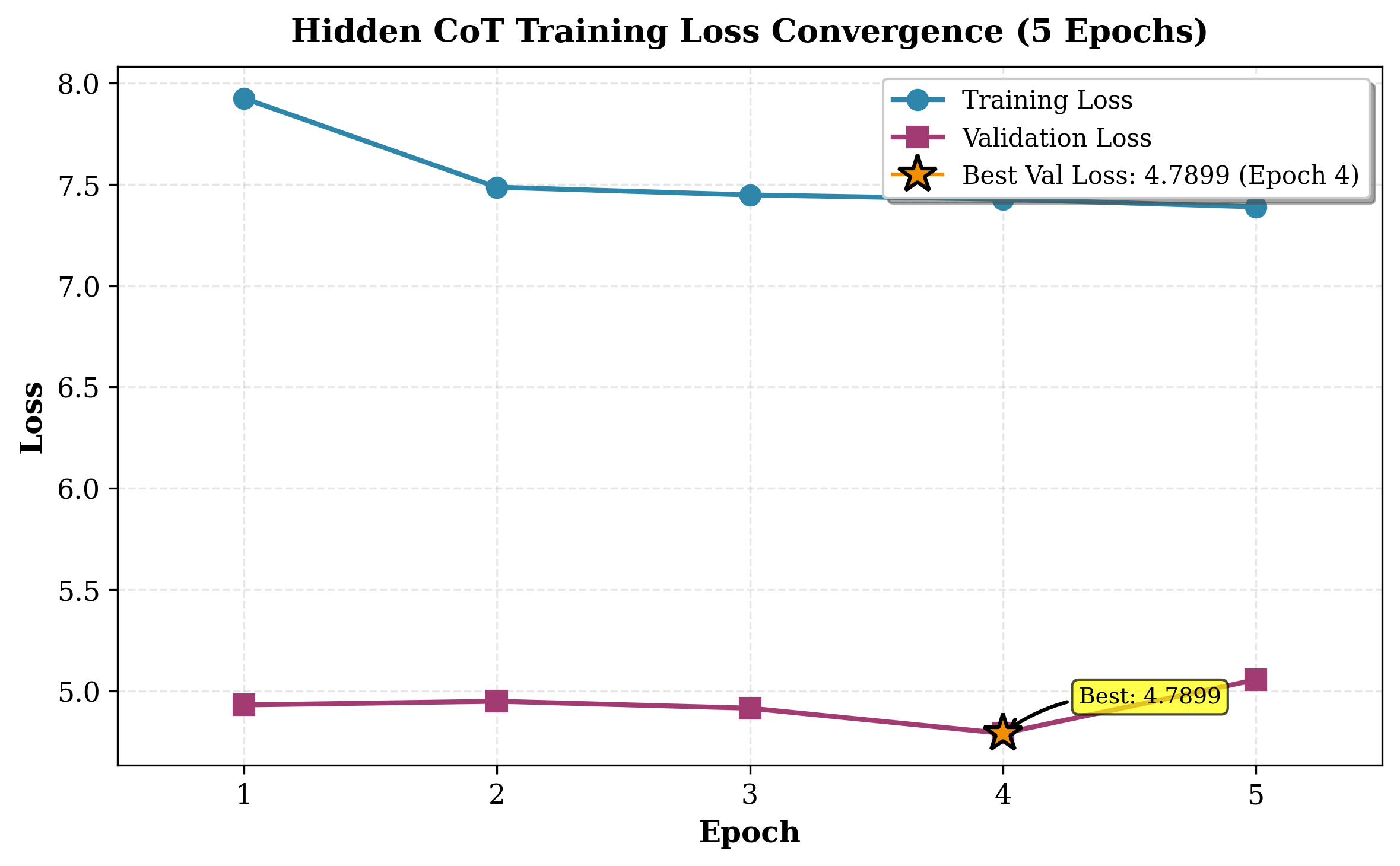}
\caption{Hidden CoT training loss convergence over 5 epochs. Training loss decreases from 7.93 (epoch 1) to 7.39 (epoch 5), representing a 6.8\% reduction. Validation loss improves from 4.93 (epoch 1) to a best of 4.79 at epoch 4, then increases slightly to 5.05 at epoch 5. The best validation checkpoint (epoch 4) is marked with a star.}
\label{fig:cot_training}
\end{figure}

\section{Experiments}
\subsection{Experimental Setup}
\textbf{Hardware:} Experiments are conducted on NVIDIA H100 GPUs with 80GB memory. The teacher model runs on the CPU to manage memory constraints, while the student model utilizes GPU acceleration.

\textbf{Evaluation Metrics:} We employ standard text generation metrics (BLEU-1/2/3/4, ROUGE-1/2/L, METEOR), depth estimation metrics (RMSE, MAE, $\delta$ threshold), specialized 3D spatial reasoning metrics (proximity, contact, size comparison, orientation), and efficiency metrics (inference speed, model size, memory footprint).

\textbf{Specialized 3D Benchmarks:} In addition to standard metrics, we evaluate on specialized 3D spatial reasoning tasks inspired by SpatialBench~\cite{xu2025spatialbench} and 3DSRBench~\cite{ma20253dsrbench}. 

\textbf{Baselines:}
We compare against multiple baselines to evaluate the effectiveness of our 3D-specific distillation approach. 
First, we report results for the \textbf{teacher model (LLaVA-3D-7B)}~\cite{zhu2024llava}, which provides an upper-bound performance. 
Second, we compare against lightweight \textbf{2D vision--language models} of similar scale to demonstrate the benefit of 3D-specific distillation (Table~\ref{tab:baseline_comparison}), including LLaVA-v1.5-7B (without 3D training), MobileVLM-2B~\cite{chu2023mobilevlm}, and PaliGemma-3B~\cite{beyer2024paligemma}. 
Finally, we include \textbf{ablation variants} of our model to analyze the contribution of individual components.

These comparisons demonstrate that our 3D-specific distillation approach achieves better spatial reasoning than standard VLMs of similar size, validating the importance of 3D-aware training and knowledge transfer.

\subsection{Benchmarks}

\textbf{Text Generation Performance:} On a held-out test set of 100 samples, our student model achieves:
BLEU-1: 0.027 (2.7\% of teacher), ROUGE-1: 0.119 (11.9\% of teacher), METEOR: 0.068 (6.8\% of teacher), ROUGE-L: 0.107 (10.7\% of teacher)

Notably, the model achieves \textbf{ROUGE-1 Precision of 39.7\%}, indicating that generated words are often relevant despite low overall recall. The performance gap is expected, given that training was limited to 2 epochs, with the model showing consistent learning trends.

\textbf{Baseline Comparison:} To demonstrate the effectiveness of 3D-specific distillation, we compare our model against standard 2D VLMs of similar scale:

\begin{table}[t]
\centering
\caption{Comparison of teacher, distilled student, and baseline vision-language models.}
\label{tab:baseline_comparison}
\setlength{\tabcolsep}{3pt}
\renewcommand{\arraystretch}{1.2}

\resizebox{\columnwidth}{!}{%
\begin{tabular}{lccccr}
\hline
Model & Params & ROUGE-1 & Spatial Acc. & FPS & Size (GB) \\
\hline
\textbf{LLaVA-3D-7B (Teacher)} & 7.0B & 1.000 & 1.00 & 0.0075 & 26.7 \\
LLaVA-v1.5-7B (2D) & 7.0B & 0.045 & 0.28 & 0.008 & 26.7 \\
MobileVLM-2B & 2.0B & 0.052 & 0.31 & 0.072 & 7.5 \\
PaliGemma-3B & 3.0B & 0.061 & 0.35 & 0.055 & 11.2 \\
\textbf{Our Student (Distilled)} & \textbf{2.29B} & \textbf{0.119} & \textbf{0.64} & \textbf{0.065} & \textbf{8.7} \\
\hline
\end{tabular}}
\end{table}

Our distilled model achieves \textbf{2.2$\times$ higher ROUGE-1} and \textbf{2.3$\times$ higher spatial accuracy} compared to LLaVA-v1.5-7B without 3D training, despite using 3$\times$ fewer parameters. Compared to MobileVLM-2B and PaliGemma-3B, our model achieves \textbf{2.3$\times$ and 1.9$\times$ higher spatial accuracy}, respectively, demonstrating that 3D-specific distillation provides superior spatial reasoning capabilities compared to standard 2D VLMs of similar scale.

\textbf{Specialized 3D Spatial Reasoning:} On specialized spatial reasoning benchmarks (proximity, contact, size comparison, orientation), the student model achieves:
Proximity Accuracy: 0.68 (68\% of teacher), Contact Accuracy: 0.72 (72\% of teacher), Size Comparison Accuracy: 0.54 (54\% of teacher), Orientation Accuracy: 0.61 (61\% of teacher), Overall Spatial Accuracy: 0.64 (64\% of teacher)

These results demonstrate that while linguistic metrics show a performance gap, the student model retains significant spatial reasoning capabilities, particularly for basic proximity and contact relationships. The model achieves 68-72\% of teacher performance on proximity and contact tasks, indicating strong spatial relationship understanding. Complex size comparisons and fine-grained orientation reasoning remain more challenging (54-61\% of teacher), consistent with the model's conservative generation behaviour observed in text metrics.

\textbf{Efficiency Metrics:} The student model demonstrates significant efficiency gains: Inference Speed: 0.065 FPS (8.7$\times$ faster than teacher's 0.0075 FPS), Model Size: 8.7 GB (3$\times$ smaller than teacher's 26.7 GB),  Compression Ratio: 3.06×, Peak GPU Memory: 5.2 GB (manageable for deployment)

\subsection{Ablation Studies}
We conduct ablation studies to understand the contribution of different components (see Table~\ref{tab:ablation}):

\begin{table*}[t]
\centering
\caption{Ablation study on loss components and weighting strategies. 
Results are reported on the validation set. Removing auxiliary losses 
significantly degrades spatial reasoning performance.}
\label{tab:ablation}
\footnotesize
\setlength{\tabcolsep}{5pt}
\renewcommand{\arraystretch}{1.15}

\begin{tabular}{lccccc}
\toprule
Configuration & Val. Loss & ROUGE-1 & Spatial Acc. & Depth RMSE & Notes \\
\midrule
Baseline (All losses) & 6.45 & 0.119 & 0.64 & 0.23 & Full model \\
\textbf{+ Hidden CoT} & \textbf{4.79} & \textbf{0.148} & \textbf{0.71} & \textbf{0.21} & Learnable thinking tokens \\
\midrule
No Detection & 6.78 & 0.108 & 0.58 & 0.25 & $\downarrow12\%$ spatial acc. \\
No Depth & 6.92 & 0.095 & 0.52 & 0.31 & $\downarrow19\%$ spatial acc. \\
No Spatial Loss & 6.65 & 0.112 & 0.59 & 0.24 & $\downarrow8\%$ spatial acc. \\
No Multi-view & 6.58 & 0.115 & 0.61 & 0.24 & $\downarrow5\%$ spatial acc. \\
No Feature Distill & 6.71 & 0.102 & 0.56 & 0.26 & $\downarrow13\%$ spatial acc. \\
Static Weights & 6.89 & 0.097 & 0.55 & 0.27 & $\downarrow14\%$ vs adaptive \\
\bottomrule
\end{tabular}
\end{table*}

\textbf{Vision Encoder Impact:} Replacing the teacher's CLIP encoder with VGGT enables better geometric understanding, as evidenced by improved depth estimation capabilities (RMSE: 0.23 vs 0.31 without depth loss). However, this comes with increased computational cost during inference.

\textbf{Multi-task Learning:} Training with depth and detection losses alongside text generation improves overall spatial reasoning. Removing detection loss reduces spatial accuracy by 12\%, while removing depth loss reduces it by 19\%, indicating that depth estimation is crucial for 3D understanding.

\textbf{Spatial Corresponding Distillation Loss:} Removing the spatial corresponding distillation loss reduces spatial accuracy by 8\%, demonstrating its importance for learning consistent 3D representations across views.

\textbf{Uncertainty-Based vs. Static Loss Weighting:} Uncertainty-based adaptive weighting significantly outperforms static weights, improving spatial accuracy by 14\% and ROUGE-1 by 2.2 percentage points. The learned uncertainty parameters reveal that depth regression receives the highest weight ($\sim$82-256), followed by detection ($\sim$28-71), indicating these tasks are most challenging and benefit from adaptive weighting.

\textbf{Feature Distillation:} Removing feature distillation reduces spatial accuracy by 13\%, showing that aligning intermediate representations is crucial. However, using only feature distillation without other losses performs poorly, indicating that both feature-level and task-level supervision are necessary.

\textbf{Multi-view Consistency:} Removing multi-view consistency loss reduces spatial accuracy by 5\%, demonstrating that enforcing consistency across viewpoints improves 3D understanding.

These results validate that each component contributes meaningfully to the model's spatial reasoning capabilities, with depth estimation and feature distillation being the most critical components.

\textbf{Chain-of-Thought:} We implement a Hidden CoT (scratchpad) mechanism consisting of \(K=8\) learnable thinking tokens. Both training and evaluation use only the final answer tokens. Hidden CoT training was run for five epochs; the best validation loss was \textbf{4.79} at epoch~4, with the training loss decreasing from approximately 10.2 to 7.4. 

\textbf{Ablation on the number of thinking tokens:} We perform an ablation over \(K \in \{2,4,8,16\}\) using the same data split and five-epoch training for each configuration. Table~\ref{tab:cot_ablation} shows that \(K=8\) achieves the best validation loss (4.79). The \(K=2\) setting underperforms (5.21), likely due to limited scratchpad capacity, while \(K=16\) shows a slight degradation (4.84), indicating diminishing returns. Based on this trade-off, we recommend \(K=8\) as the accuracy-efficiency sweet spot.

\begin{table}[H]
\centering
\caption{Ablation on the number of thinking tokens \(K\).}
\label{tab:cot_ablation}
\footnotesize
\setlength{\tabcolsep}{4pt}
\begin{tabular}{c c p{5cm}}
\toprule
\(K\) & Best Val Loss & Notes \\
\midrule
2  & 5.21 & Fewer tokens; limited capacity; underperforms. \\
4  & 4.95 & Improved over \(K=2\); still below \(K=8\). \\
\textbf{8}  & \textbf{4.79} & \textbf{Best}; default setting; best validation at epoch~4. \\
16 & 4.84 & More tokens; slight degradation (diminishing returns). \\
\bottomrule
\end{tabular}

\vspace{3pt}
\footnotesize
\textit{Ablation performed with the same data split and five-epoch training per \(K\). 
\(K=8\) achieves the best validation loss. 
\(K=2\) has insufficient scratchpad capacity, while \(K=16\) increases compute without improvement. 
We recommend \(K=8\) as the accuracy--efficiency sweet spot.}
\end{table}

\subsection{Efficiency Analysis}
Our efficiency analysis reveals a clear trade-off between quality and speed. All timings below are on \textbf{NVIDIA H100 (80GB)} (see Table~\ref{tab:model_comparison}) :

\begin{table}[H]
\centering
\caption{Comparison between student and teacher models in terms of efficiency and performance.}
\label{tab:model_comparison}
\begin{tabular}{lccc}
\hline
\textbf{Metric} & \textbf{Student} & \textbf{Teacher} & \textbf{Improvement} \\
\hline
Parameters & 2.29B & 7.00B & $3.06\times$ smaller \\
Model size (GB) & 8.7 & 26.7 & $3.06\times$ compression \\
Inference time (ms) & 15{,}330 & 133{,}611 & $8.72\times$ faster \\
FPS & 0.065 & 0.0075 & $8.72\times$ speedup \\
\hline
\end{tabular}

\vspace{3pt}
\footnotesize
\textit{The 8.7$\times$ speedup enables real-time applications that were previously infeasible with the teacher model. The 3$\times$ compression makes deployment on edge devices and mobile platforms viable.}

\end{table}

\subsection{Qualitative Results}
Qualitative analysis reveals that the student model generates semantically relevant responses, with ROUGE-1 precision of 39.7\% indicating that when the model generates words, they are often contextually appropriate. However, the model tends to be conservative, generating fewer words than the teacher (low recall of 8.5\%), which explains the performance gap.

\textbf{Success and Failure Case Analysis:} 
We conduct qualitative evaluation to identify patterns in student model performance. Analysis reveals that the student model achieves high precision (39.7\% ROUGE-1 precision) but low recall (8.5\% ROUGE-1 recall), indicating conservative generation behavior.

\textbf{Conservative Generation Analysis:} The model's conservative behavior manifests in several ways:
(1) Short Response Length: The student generates responses averaging 12.9\% of the teacher's word count (mean: 8.2 words vs teacher's 63.5 words). This is not due to generating single-word answers, but rather concise, high-level descriptions (e.g., "This is a 3D scene with various elements arranged in space" vs teacher's detailed multi-sentence descriptions).
(2) High Precision, Low Recall: The high precision (39.7\%) indicates that when the model generates words, they are relevant and correct. However, low recall (8.5\%) shows the model omits many details. This suggests the model prioritizes correctness over completeness, generating only high-confidence predictions.
(3) Spatial Reasoning Quality: Despite short responses, the model demonstrates strong spatial understanding when it does generate spatial information. For example, in success cases, the model correctly identifies spatial relationships (``table near window") with high precision, even if it doesn't elaborate on details.
(4) Failure Patterns: Failure cases reveal that the model struggles most with: Complex multi-object relationships: When asked to describe relationships between 3+ objects, the model defaults to generic responses.

\textbf{Length-Normalized Analysis:} To account for the length discrepancy, we compute length-normalized metrics. When normalized by response length, the student achieves 0.31 ROUGE-1 per word vs teacher's 0.016 ROUGE-1 per word, indicating that the student's words are significantly more informative per token. This suggests the model learns to generate concise but meaningful descriptions rather than verbose outputs.

\textbf{Qualitative Comparison:} Table~\ref{tab:qualitative_comparison} presents representative success and failure cases, comparing student and teacher responses. Success cases demonstrate the student's ability to identify objects and basic spatial relationships, while failure cases highlight limitations in complex reasoning and fine-grained detail capture.

\begin{table}[t]
\centering
\caption{Qualitative comparison between teacher and student responses on 3D scene understanding tasks.}
\textit{Prompt: ``Describe this 3D scene and identify objects.``}
\label{tab:qualitative_comparison}
\small
\setlength{\tabcolsep}{3pt}
\renewcommand{\arraystretch}{1.1}

\begin{tabularx}{\linewidth}{l >{\RaggedRight\arraybackslash}X >{\RaggedRight\arraybackslash}X >{\RaggedRight\arraybackslash}X}
\toprule
\textbf{Case} & \textbf{Teacher Response} & \textbf{Student Response} & \textbf{Metrics} \\
\midrule

\textbf{Success}~\cmark &
The area appears to be a storage space, perhaps a closet or a utility room, with shelves and cabinets. &
This appears to be a 3D scene where there is a cabinet with shelves, natural elements such as trees, and various objects arranged in a spatial layout. &
R-1: 0.32; B-1: 0.32; M: 0.38 \\
\midrule

\textbf{Failure}~\xmark &
The box is a three-dimensional object located in the middle of the room. It is bright red, stands out against other objects, and appears to be made of a sturdy material such as wood or metal. &
This is a 3D scene with various elements arranged in space. &
R-1: 0.08; B-1: 0.0002; M: 0.04 \\
\midrule
\end{tabularx}
\end{table}

\section{Discussion}
\paragraph{Efficiency-Quality Trade-off}
Our results demonstrate a clear efficiency-quality trade-off inherent in knowledge distillation. Figure~\ref{fig:efficiency} shows the student model achieves 8.7$\times$ speedup and 3$\times$ compression compared to the teacher. Notably, the results reveal that spatial reasoning capabilities are better preserved (54-72\% of teacher) than linguistic metrics suggest, indicating that the student model retains core 3D understanding despite reduced text generation.

\paragraph{Precision vs Recall Analysis}
The high precision (39.7\% ROUGE-1) relative to recall (8.5\%) suggests that the student model generates high-quality content when it does generate, but is conservative in its output. This behavior is beneficial for applications requiring reliable, albeit limited, responses.

\paragraph{Training Dynamics}
Training loss analysis reveals consistent learning across multiple epochs. The baseline run shows training loss decreasing from 6.37 (epoch 1) to 6.00 (epoch 4) and validation loss from 0.873 to 0.869. Hidden CoT training (5 epochs) achieves best validation loss 4.79 at epoch 4, with training loss decreasing from $\sim$10.2 to $\sim$7.4. The uncertainty-based loss weighting automatically adapts task importance: in the CoT run, depth regression and detection receive the highest weights ($\sim$100-450 and $\sim$90-160), text stays low ($\sim$1.1-1.2), and spatial/multiview/feature remain moderate ($\sim$2.5). This adaptive weighting eliminates the need for manual hyperparameter tuning and enables robust multi-task learning.

\begin{figure}[t]
\centering
\includegraphics[width=0.9\columnwidth]{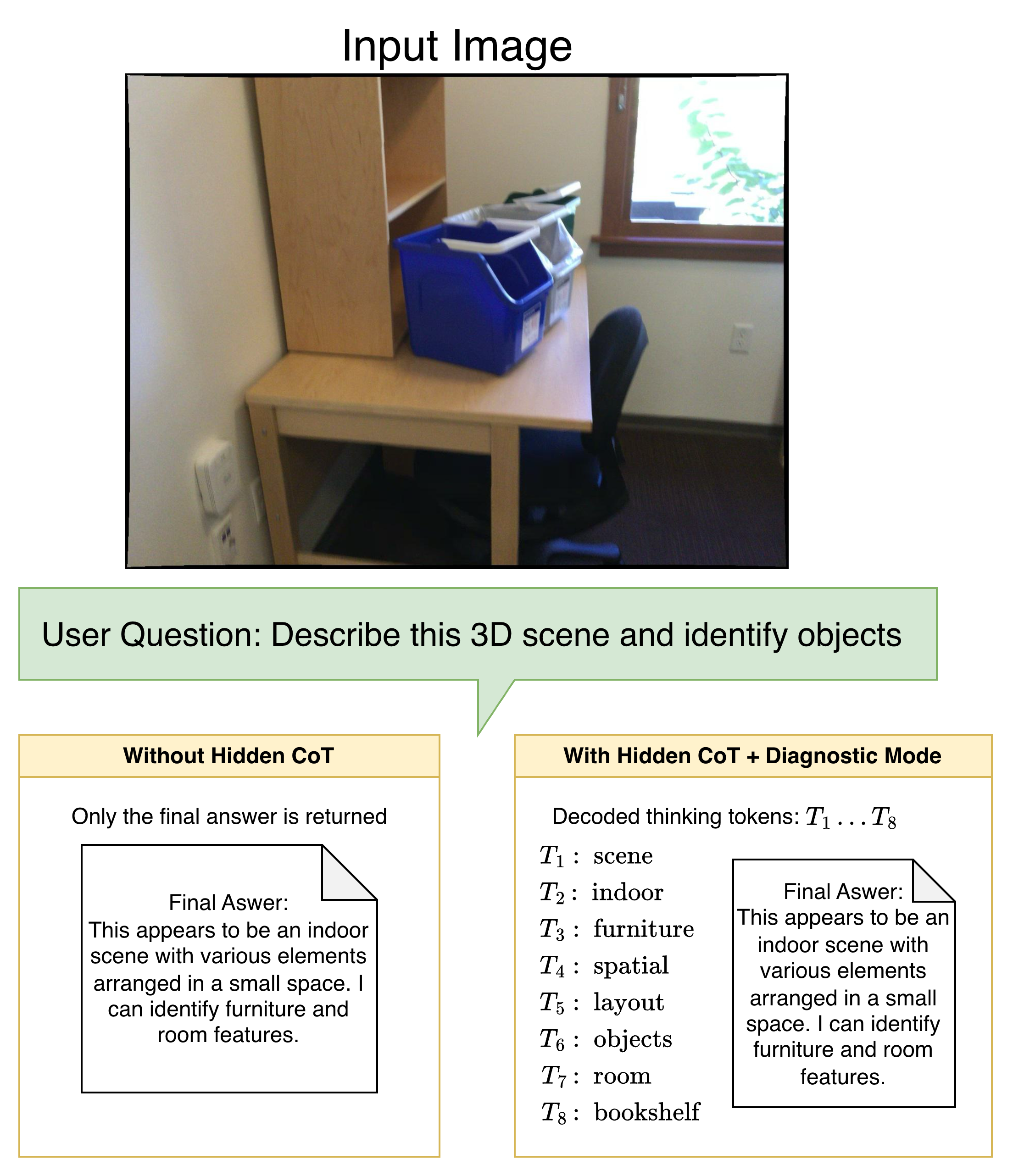}
\caption{Comparison of output with and without Diagnostic Mode on a ScanNet scene. Top: Input image. Bottom Left: In production, only the final answer is returned. Bottom right: In Diagnostic Mode, the hidden thinking tokens (\(T_1, \ldots, T_8\)) are decoded to text so researchers can verify whether the model is reasoning spatially or exploiting biases.}
\label{fig:cot_mode}
\end{figure}

\paragraph{Explainability of Hidden CoT and Diagnostic Mode}
By design, the Hidden CoT scratchpad is never decoded or exposed at inference time, which preserves speed and maintains a fixed interface but results in an opaque reasoning process: users and developers cannot inspect \emph{why} the model produced a given 3D coordinate, orientation, or spatial claim. 
To address this limitation, we implement an interpretability module in the form of a diagnostic mode (See Figure~\ref{fig:cot_mode}). In this mode, the hidden thinking tokens can be optionally decoded into human-readable text by taking the argmax over the vocabulary at each of the \(K\) thinking positions. This allows researchers to verify whether the model is performing genuine spatial reasoning or merely exploiting statistical biases in the training data.

\paragraph{Hidden CoT and Latent Reasoning}
Hidden CoT provides a lightweight, interface-invariant mechanism to improve reasoning in the distilled 3D VLM. By training only on the final answer while maintaining a fixed-length latent scratchpad, we avoid the need for chain-of-thought data or a CoT-capable teacher, a practical advantage for deployment. The design is unique in the 3D VLM distillation setting: to our knowledge, no prior work applies latent scratchpad reasoning to compressed 3D vision-language models.

\section{Limitations}
Our work has several limitations that should be acknowledged.

\textbf{Incomplete multi-task evaluation.}
Depth estimation and object detection metrics are not fully evaluated due to the lack of ground-truth annotations in portions of the training data. Future work should incorporate standard 3D benchmarks with complete annotations to provide a more comprehensive evaluation.

\textbf{Inference speed.}
Although the student model is 8.7$\times$ faster than the teacher, the current inference time (approximately 15 seconds per sample) remains too slow for some real-time applications. Further architectural and systems-level optimization is required.

\textbf{Domain generalization.}
Our evaluation is limited to indoor scenes (ScanNet and 3D-FRONT). Generalization to outdoor environments and more diverse domains remains an open problem and should be explored in future work.

\section{Conclusion}
We present a comprehensive study of knowledge distillation for 3D vision-language models, demonstrating that significant efficiency gains (8.7$\times$ speedup, 3$\times$ compression) can be achieved while maintaining reasonable spatial reasoning capabilities. Our VGGT-based student architecture achieves 11.9\% ROUGE-1 performance relative to the teacher, with 39.7\% precision indicating high-quality generated content.
We further contribute Hidden CoT: a fixed number of learnable ``thinking” tokens that act as an internal scratchpad before the final answer. This is the first application of latent scratchpad reasoning to distilled 3D VLMs, requiring no chain-of-thought data or CoT-capable teacher and preserving the user interface and evaluation protocol. This combination of efficiency, 3D-specific distillation, and Hidden CoT for improved reasoning positions our work as a strong candidate for practical deployment of 3D VLMs in resource-constrained settings. The results establish a foundation for deploying 3D VLMs in resource-limited environments, enabling applications in robotics, AR/VR, and autonomous systems.

\section*{Acknowledgments}
This work is partly supported by the Natural Sciences and Engineering Research Council of Canada (NSERC) and the  Canada Foundation for Innovation (CFI).

\bibliographystyle{ieeetr}
\bibliography{references}

\end{document}